\documentclass[letterpaper, 10 pt, conference]{ieeeconf}  

\IEEEoverridecommandlockouts                              

\usepackage{graphics} 
\usepackage{epsfig} 
\usepackage{mathptmx} 
\usepackage{times} 
\usepackage{amsmath} 
\usepackage{amssymb}  

\usepackage{booktabs}
\usepackage{multirow}
\usepackage{makecell}
\usepackage{caption}
\usepackage{subfig}

\def\eg{\emph{e.g.}} 

\def\ie{\emph{i.e.}}

\def\etal{\emph{et al.}}

\newcommand{\ts}{\textsuperscript}

\title{\LARGE \bf
NeRF-Enhanced Outpainting for Faithful Field-of-View Extrapolation
}

\author{Rui Yu$^{1}$$^\dagger$\thanks{$^\dagger$Equal contribution to this work.}, Jiachen Liu$^{2}$$^\dagger$, Zihan Zhou$^{3}$, Sharon X. Huang$^{2}$
\thanks{$^{1}$Rui Yu is with the Department of Computer Science and Engineering, University of Louisville, Louisville, KY 40208, USA {\tt\small rui.yu@louisville.edu}}%
\thanks{$^{2}$Jiachen Liu and Sharon X. Huang are with College of Information Sciences and Technology, Pennsylvania State University, University Park, PA 16802, USA {\tt\small \{jzl6493, suh972\}@psu.edu}}%
\thanks{$^{3}$Zihan Zhou is with Manycore Tech Inc., Hangzhou, Zhejiang, China {\tt\small shuer@qunhemail.com}}%
}

\begin{document}

\maketitle
\thispagestyle{empty}
\pagestyle{empty}

\begin{abstract}

In various applications, such as robotic navigation and remote visual assistance, expanding the field of view~(FOV) of the camera proves beneficial for enhancing environmental perception. Unlike image outpainting techniques aimed solely at generating aesthetically pleasing visuals, these applications demand an extended view that faithfully represents the scene. To achieve this, we formulate a new problem of faithful FOV extrapolation that utilizes a set of pre-captured images as prior knowledge of the scene. To address this problem, we present a simple yet effective solution called NeRF-Enhanced Outpainting (NEO) that uses extended-FOV images generated through NeRF to train a scene-specific image outpainting model. To assess the performance of NEO, we conduct comprehensive evaluations on three photorealistic datasets and one real-world dataset. Extensive experiments on the benchmark datasets showcase the robustness and potential of our method in addressing this challenge. We believe our work lays a strong foundation for future exploration within the research community.

\end{abstract}

\section{INTRODUCTION}
\label{sec:intro}

\begin{figure*}[t]
\centering
\includegraphics[width=0.75\linewidth]{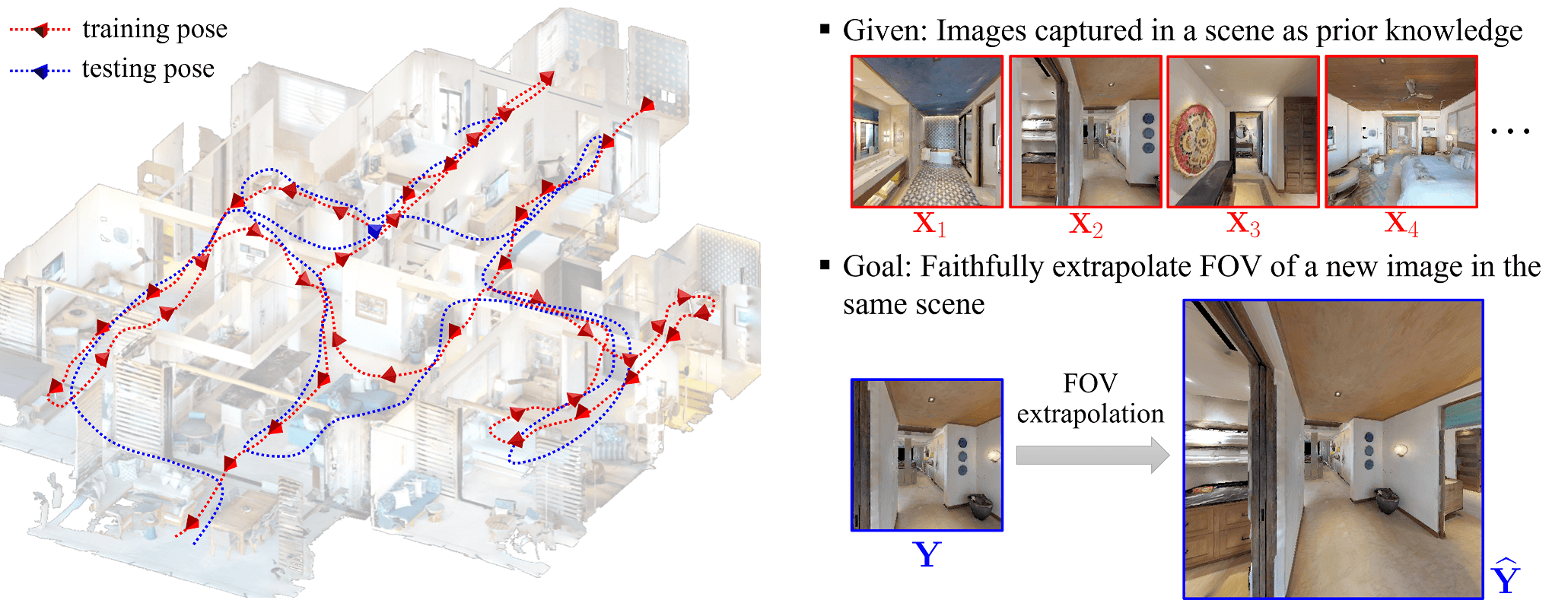} 
\caption{Problem formulation of faithful FOV extrapolation. A collection of training images $\mathbf{X}_i$ (red box) taken within a specific scene serves as prior knowledge. In the testing phase, our objective is to faithfully extrapolate the FOV of newly captured image $\mathbf{Y}$ (blue box) to a specified FOV by leveraging the prior knowledge of the scene.}
\label{fig:problem_form}
\vspace{-2ex}
\end{figure*}

The field of view (FOV) of a camera plays a pivotal role in the performance of vision-based navigation~\cite{DBLP:conf/icra/ZhuMKLGFF17,DBLP:conf/cvpr/AndersonWTB0S0G18}. A larger FOV enables robots to perceive more spatial elements and layouts (\eg, obstacles, doorways, etc.). This expanded perspective empowers them to make more informed and strategic decisions when planning their paths. A larger FOV also offers substantial benefits for remote sighted agents (RSAs) tasked with assisting visually impaired individuals in navigation~\cite{DBLP:conf/iui/LeeYXBC22,DBLP:conf/ACMdis/XieYLLBC22}. In light of this motivation, our work delves into the challenge of FOV extrapolation. Our goal is to enable robots and remote agents to perceive scene content beyond the immediate camera FOV, thereby enhancing their situational awareness.

In the computer vision domain, our task is closely related to image outpainting~\cite{DBLP:conf/iccv/YangDLYY19,DBLP:conf/cvpr/LiCX22,DBLP:conf/cvpr/ChengLLRT022,DBLP:conf/eccv/YaoG0S0H22}, also referred to as image extrapolation~\cite{DBLP:conf/cvpr/WangTSJ19,DBLP:conf/eccv/GuoLZCSGZZ20,DBLP:conf/iccv/KhuranaDBMSK21} or image extension~\cite{DBLP:conf/iccv/KrishnanTSMLBF19}, which aims to extend the image boundaries with semantically consistent and visually appealing contents. The extrapolated image hallucinates visually plausible scene contents beyond the original FOV, delivering immersive viewing experience for applications such as virtual reality. The hallucination ability is acquired by, for example, training deep learning models on large-scale generic datasets~\cite{DBLP:journals/pami/ZhouLKO018} of realistic images. However, image outpainting models cannot be applied to our problem because navigational applications necessitate the extended portions of the image maintain fidelity and coherence with the actual scene.

We define the problem of \textit{faithful FOV extrapolation} as follows. As illustrated in Fig.~\ref{fig:problem_form} (right), a collection of training images that were captured in a given scene~(\eg, images shown inside red boxes) serves as prior knowledge. We assume that the camera pose corresponding to each training image can be obtained, for example, via structure from motion~(SfM)~\cite{lindenberger2021pixel, DBLP:conf/cvpr/SarlinULGTLPLHK21, schonberger2016structure} methods. During the testing phase, our objective is to faithfully extrapolate the FOV of any newly captured image in the same scene to a specified FOV based on the prior knowledge of the scene (see example images inside blue boxes).

It is possible to adapt existing computer vision techniques, namely image stitching and video expansion, to tackle the faithful FOV extrapolation problem. Image stitching~\cite{DBLP:journals/ftcgv/Szeliski06,DBLP:journals/tmm/LiXW18,DBLP:journals/tip/Liao020,DBLP:conf/cvpr/LeeS20} entails aligning overlapping portions of multiple images taken from various angles, whereas video expansion~\cite{wang2018biggerselfie}, or video extrapolation~\cite{DBLP:journals/tog/LeeLKKN19,ma2021fov}, leverages adjacent frames to extend the FOV of a specific frame. Both methods require precise warping of source images to blend seamlessly with the target image. However, they often yield irregularly shaped non-overlapping areas, imposing limitations on the extent of FOV expansion. Therefore, these methods are not well suited for our goal, as the faithful FOV extrapolation task demands expanding the view to a desired rectangular size.

Given that there has been very limited prior research addressing the challenge of faithful FOV extrapolation for navigation, we propose a simple yet effective method called \textit{NeRF-Enhanced Outpainting (NEO)}. Our method involves first training a neural radiance fields~\cite{DBLP:conf/eccv/MildenhallSTBRN20} (NeRF) model using training images of original FOV. We then densely sample a substantial number of camera poses within the same scene and, for each sampled pose, the trained NeRF model is applied to render an image with an expanded FOV. Finally, we leverage these rendered images to train an image outpainting model, which is subsequently employed to extrapolate the FOV of input images during the inference phase. We validate the proposed method on three photorealistic datasets and one real dataset. The NEO method excels at producing high-quality extrapolations tailored to the specified FOV and consistently surpasses the performance of three baseline methods.

In summary, the contributions of this work are as follows. (1) We introduce a novel problem, namely faithful FOV extrapolation for navigation, which has been relatively underexplored in existing literature. (2) We propose the NeRF-Enhanced Outpainting (NEO) pipeline as a solution for faithful FOV extrapolation. (3) Comprehensive empirical investigations on both photorealistic and real-world datasets consistently validate the effectiveness of the proposed NEO method compared to the baseline counterparts.

\section{Related Work}
\label{sec:related_work}

\subsection{Image Outpainting}

Image outpainting, often referred to as image extrapolation or extension, is a task that seeks to expand image boundaries while maintaining semantically coherent content. Typically, the ability to infer such contents is acquired through learning from large-scale datasets of real images.
Image outpainting approaches can be broadly categorized as either non-parametric or parametric.
Non-parametric methods~\cite{DBLP:conf/iccv/EfrosL99,DBLP:conf/siggraph/EfrosF01} are restricted to basic pattern outpainting, and they become increasingly fragile as the extrapolation range grows.
The emergence of GAN-based models has resulted in significant advancements in image outpainting. Some notable works~\cite{DBLP:conf/cvpr/Yeh0LSHD17,DBLP:conf/cvpr/WangTSJ19,DBLP:conf/iccv/KrishnanTSMLBF19,DBLP:conf/iccv/YangDLYY19,DBLP:conf/eccv/GuoLZCSGZZ20} utilize a single GAN model for extrapolating the input image.
More recently, Khurana \etal~\cite{DBLP:conf/iccv/KhuranaDBMSK21} propose an image outpainting framework that extends the image within the semantic label space, thereby produce new objects within the extrapolated area.
Li \etal~\cite{DBLP:conf/cvpr/LiCX22} introduce CTO-GAN, which deduces the potential semantic layout based on foreground elements and subsequently generates the corresponding background content with the guidance of the predicted semantics.
Yao \etal~\cite{DBLP:conf/eccv/YaoG0S0H22} formulate this problem as a sequence-to-sequence autoregression task based on image patches, and present a query-based encoder-decoder transformer model to perform extrapolation.
In addition to dedicated outpainting methods, certain image inpainting models~\cite{DBLP:conf/iclr/ZhaoCSDLCX21,DBLP:conf/wacv/SuvorovLMRASKGP22,DBLP:conf/cvpr/LiLZQWJ22}, which have the capability to fill large masks, can also be adapted for image outpainting. Inspired by the pioneering work~\cite{ho2020denoising} on diffusion models, there have been endeavors~\cite{saharia2022palette, rombach2022high, lugmayr2022repaint, caidiffdreamer} that tackle the image outpainting problem via a diffusion-then-denoising process. 
One limitation of these pretrained image outpainting models is that the extrapolated parts lack geometric consistency and interpretability for a specific scene. This characteristic renders them unsuitable for real-world application scenarios such as navigation. Therefore, we propose a new problem, faithful FOV extrapolation, the solution of which enables and facilitates navigational applications by ensuring that the extrapolated content remains faithful and relevant to the scene at hand.

\subsection{NeRF and Data Augmentation}

Neural radiance fields (NeRF)~\cite{DBLP:conf/eccv/MildenhallSTBRN20} enables novel view synthesis by representing the density and color of 3D spatial points of a specific scene through a neural network.
With a novel camera pose as input, NeRF can render an image of specified FOV by performing ray marching from the camera's central viewpoint, querying the corresponding color and density fields and conducting volume rendering. Follow-up works on NeRF have explored improving generalizability~\cite{schwarz2020graf, yu2021pixelnerf}, scene editing~\cite{liu2021editing, zhang2021editable, wang2022clip}, neural scene reconstruction~\cite{wang2021neus, yariv2021volume}, training and inference acceleration~\cite{DBLP:journals/tog/MullerESK22, DBLP:conf/eccv/ChenXGYS22}, among others. Unlike other image outpainting techniques which rely on scene distribution priors to extrapolate large FoV, NeRF has its unique advantage in that it implicitly encodes the entire 3D scene, enabling the rendering of novel views in a manner that is both geometrically and semantically coherent. This positions NeRF as a potential approach to achieving faithful FOV extrapolation. 
Furthermore, NeRF has been utilized as a data augmentation tool to generate synthetic images for training deep neural networks. 
Moreau \etal~\cite{moreau2022lens} employ NeRF model to create a fresh dataset of synthetic images for training a camera pose regression model.
Ge \etal~\cite{ge2022neural} propose an online data augmentation pipeline based on NeRF synthesis for real-world object detection.
In this paper, we present a NeRF-enhanced outpainting pipeline that leverages NeRF to generate sufficient synthetic images for training an FOV extrapolation model.

\begin{figure*}[t]
\centering
\includegraphics[width=0.95\linewidth]{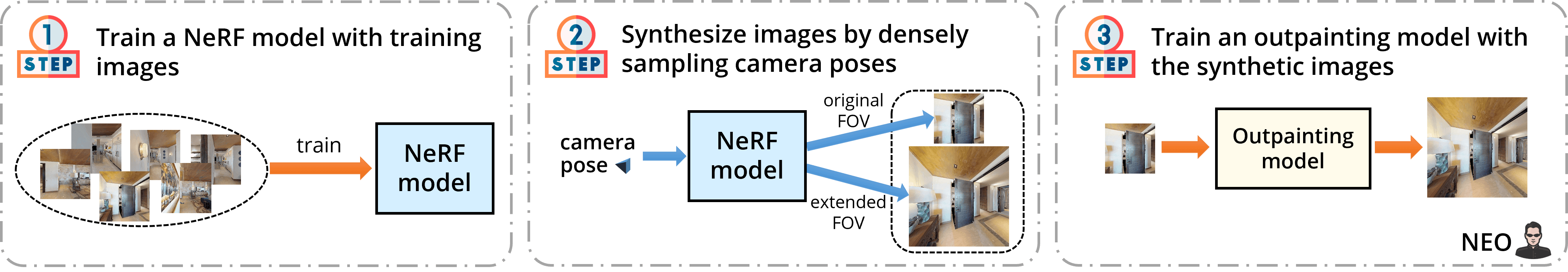} 
\caption{\textbf{NEO pipeline:} (1) training a NeRF model with captured training images of original small FOV; (2) using the trained NeRF to synthesize images of extended FOV by densely sampling camera poses in the scene; (3) training an outpainting model with the synthetic images of extended FOV. During inference, we use the trained outpainting model for faithful FOV extrapolation.}
\label{fig:neo_pipeline}
\end{figure*}

\section{Proposed method}
\label{sec:method}

\subsection{Problem Formulation}
\label{sec:prob_form}


As shown in Fig.~\ref{fig:problem_form}, a camera captured $N$ training images $\left\{ \mathbf{X}_i\in \mathbb{R}^{h\times w\times 3}|i=1,2,\dots,N \right\}$ in a specific scene, which may have been taken sparsely. The FOV of each image is $(\alpha_x, \alpha_y)$, indicating the horizontal and vertical FOV angles, respectively.
We assume that the camera pose $\mathbf{P}_i \in \mathbf{SE}(3)$ for each training image $\mathbf{X}_i$ can be acquired through structure from motion (SfM) pipelines such as COLMAP~\cite{schonberger2016structure}.
During testing, we seek to extrapolate a testing image $\mathbf{Y}\in \mathbb{R}^{h\times w\times 3}$ with the same FOV $(\alpha_x, \alpha_y)$ captured in the same scene to a new image $\widehat{\mathbf{Y}}\in \mathbb{R}^{H\times W\times 3}$ with larger FOV $(\gamma_x, \gamma_y)$ while keeping the camera's focal length constant.
The extrapolated portions of the image must be consistent with the real scene.


\subsection{NeRF-Enhanced Outpainting (NEO)}


To address this problem, we propose a simple yet effective method, dubbed NeRF-Enhanced Outpainting (NEO) with three steps in the training stage, as illustrated in Fig.~\ref{fig:neo_pipeline}.

\smallskip
\noindent{\bf Step 1: Training a NeRF model.} \hspace{0.5mm}
We start by training a NeRF model using training images $\left\{ \mathbf{X}_i | i=1,2,\dots,N \right\}$.
NeRF learns an implicit representation of the given 3D scene with a multilayer perceptron (MLP). With the trained NeRF, we can emit a ray from any direction and sample points on the ray to obtain their density and radiance, then render a novel view through volume rendering.
Therefore, NeRF can generate an image of a specified FOV with a new camera pose. Using the trained NeRF, we can render an arbitrary number of images with specified poses and desired FOV.

\smallskip
\noindent{\bf Step 2: Synthesizing images.} \hspace{0.5mm}
Next, we sample a multitude of new camera poses within the scene. For each pose, we leverage the trained NeRF model to render a pair of images with FOVs of $(\alpha_x, \alpha_y)$ and $(\gamma_x, \gamma_y)$, respectively. 
We sample new camera poses by ensuring that they cover all walkable areas in the training trajectories. Moreover, we manage to sample poses with different degrees of freedom (DoF) following their DoF distribution in the specific environment. The details will be presented in Sec.~\ref{sec:experiments}.

\smallskip
\noindent{\bf Step 3: Training an outpainting model.} \hspace{0.5mm}
Finally, we utilize the NeRF-rendered images as training data to train an image outpainting model, which takes the small-FOV images as input and extrapolates the large-FOV ones. 
Various image outpainting models~\cite{DBLP:conf/cvpr/ChengLLRT022} can be employed in this step.
Image inpainting models~\cite{DBLP:conf/cvpr/LiLZQWJ22, DBLP:conf/wacv/SuvorovLMRASKGP22} that fill in large empty spaces can also be applied to outpainting. However, it is worth noting that some outpainting or inpainting models are designed to generate diverse results with randomness. This property does not align with the objective of faithful FOV extrapolation. Thus, we modify such models during our implementation for the purpose of training an outpainting model that performs deterministic extrapolations faithful to the environment.  

During inference, a real small-FOV image is given to the trained outpainting model as input and the model performs faithful FOV extrapolation to obtain the large-FOV image. 

\subsection{Discussions}


\textit{1)\label{discussion1} Why not directly train an outpainting model using the training images $\left\{ \mathbf{X}_i | i=1,2,\dots,N \right\}$?}

The outpainting model takes $\mathbf{X}_i\in \mathbb{R}^{h\times w\times 3}$ as input and produces $\widehat{\mathbf{X}}_i\in \mathbb{R}^{H\times W\times 3}$ as output. However, there are no $\widehat{\mathbf{X}}_i$ in the training data. To address this issue, we could simply resize $\mathbf{X}_i\in \mathbb{R}^{h\times w\times 3}$ to $\mathbf{X}'_i\in \mathbb{R}^{H\times W\times 3}$ and then crop the central part $\mathbf{X}''_i\in \mathbb{R}^{h\times w\times 3}$ to use as input.
Since the training is simply achieved through resizing and cropping the original small-FOV training images, we refer to this method as~\textit{naive outpainting}.
However, this approach encounters two main challenges. First, the quantity and coverage of the training data prove inadequate for hallucinating from a new viewpoint. Second, cropping the central part reduces the FOV of the training image, leading to a mismatch of the FOVs during training and testing stages. 

Our proposed NEO pipeline addresses the above two challenges by leveraging NeRF-based synthesis. First, there is no longer a concern about limited amount of training data, as
NEO can theoretically generate an unlimited number of synthetic images by sampling arbitrary camera poses across walkable areas.  Second, the issue of training-testing FOV mismatch is resolved, because
the NEO pipeline trains the outpainting model using synthetic images with the same FOV as the target of the testing stage.
The underlying principle of NEO is that the resulting outpainting model learns to extrapolate faithfully in the given scene by processing an extensive volume of training images that comprehensively cover the entire scene.

\textit{2) Why not directly synthesize the target images using the trained NeRF model?}

NeRF can generate a specific extended-FOV image given a camera pose. However, during the testing phase, the camera pose of the testing image $\mathbf{Y}$ is unknown.
We could employ camera relocalization (also known as visual localization) paradigms~\cite{DBLP:conf/cvpr/SarlinULGTLPLHK21, schonberger2016structure} to estimate the camera pose of $\mathbf{Y}$.
Yet a main issue with combining relocalization and NeRF is that, 
the estimated pose may not be precise enough due to errors in feature detection, matching, as well as perspective-n-point~(PnP)~\cite{fischler1981random} estimation. Consequently, the resulting extended-FOV image rendered by NeRF may not be aligned well with the testing image $\mathbf{Y}$.
In contrast, the NEO pipeline circumvents the need for highly accurate estimation of the camera poses for testing images. Instead, it benefits from NeRF by training an outpainting model with the underlying scene priors from NeRF renderings.


\section{Experiments}
\label{sec:experiments}

\begin{table*}[t]
\centering
\small
\caption{\textbf{Quantitative evaluation on four datasets.} The backbone for computing LPIPS metrics is VGG network.}
\label{table:quant_results}
\captionsetup{singlelinecheck = false}
\resizebox{0.95\linewidth}{!}{
\begin{tabular}{ l  c c c  c c c  c c c  c c c} 
\toprule
\multirowcell{2}{Method} & \multicolumn{3}{c}{\textbf{Replica~(photorealistic)}} & \multicolumn{3}{c}{\textbf{Gibson~(photorealistic)}} & \multicolumn{3}{c}{\textbf{HM3D~(photorealistic)}} & \multicolumn{3}{c}{\textbf{ScanNet~(real)}} \\ 
\cmidrule(lr){2-4}
\cmidrule(lr){5-7}
\cmidrule(lr){8-10}
\cmidrule(lr){11-13}

& {PSNR $\uparrow$} & {SSIM $\uparrow$} & {LPIPS $\downarrow$} & {PSNR $\uparrow$} & {SSIM $\uparrow$} & {LPIPS $\downarrow$} & {PSNR $\uparrow$} & {SSIM $\uparrow$} & {LPIPS $\downarrow$} & {PSNR $\uparrow$} & {SSIM $\uparrow$} & {LPIPS $\downarrow$} \\
\midrule
{(B1) Naive Outpainting~\cite{DBLP:conf/cvpr/LiLZQWJ22}} & 20.59 & 0.781 & 0.348 & 18.05 & 0.705 & 0.404 & 17.92 & 0.630 & 0.427 & 19.98 & 0.755 & 0.188 \\
{(B2) Warping \& Fusion~\cite{DBLP:conf/iccv/LiuL0CH21}} & 19.03 & 0.745 & 0.397 & 16.61 & 0.688 & 0.496 & 15.94 & 0.582 & 0.512 & 21.02 & 0.755 & 0.238 \\
{(B3) Relocalized NeRF~\cite{DBLP:conf/cvpr/0004SC22, dai2023hybrid, schonberger2016structure}} & 16.78 & 0.724 & 0.386 & 14.90 & 0.641 & 0.474 & 14.43 & 0.602 & 0.484 & 18.88 & 0.695 & 0.188 \\
\midrule
{Oracle NeRF~\cite{DBLP:conf/cvpr/0004SC22, dai2023hybrid}} & 32.90 & 0.936 & 0.174 & 32.16 & 0.928 & 0.188 & 27.03 & 0.824 & 0.299 & 23.80 & 0.805 & 0.108 \\
\midrule
{NEO} & 25.94 & 0.868 & 0.217 & 23.53 & 0.822 & 0.263 & 21.54 & 0.731 & 0.338 & 22.40 & 0.793 & 0.168 \\
\bottomrule
\end{tabular}
}
\end{table*}

\begin{figure*}[t]
\centering
\includegraphics[width=0.92\linewidth]{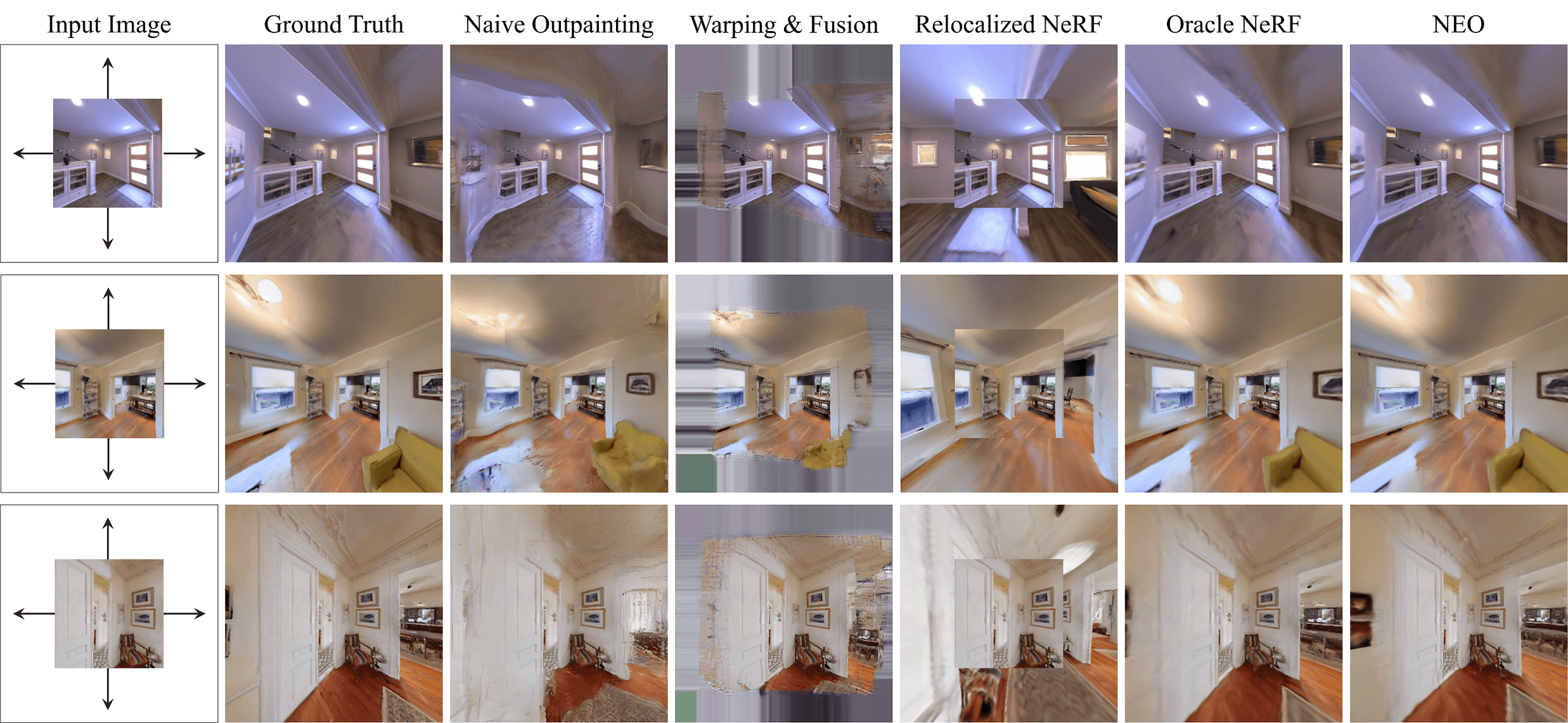} 
\caption{Qualitative results on three photorealistic datasets: \textbf{Replica} (1\ts{st} row), \textbf{Gibson} (2\ts{nd} row), and \textbf{HM3D} (3\ts{rd} row).}
\label{fig:vis_result_synthetic}
\end{figure*}

\begin{figure*}[t]
\centering
\includegraphics[width=0.92\linewidth]{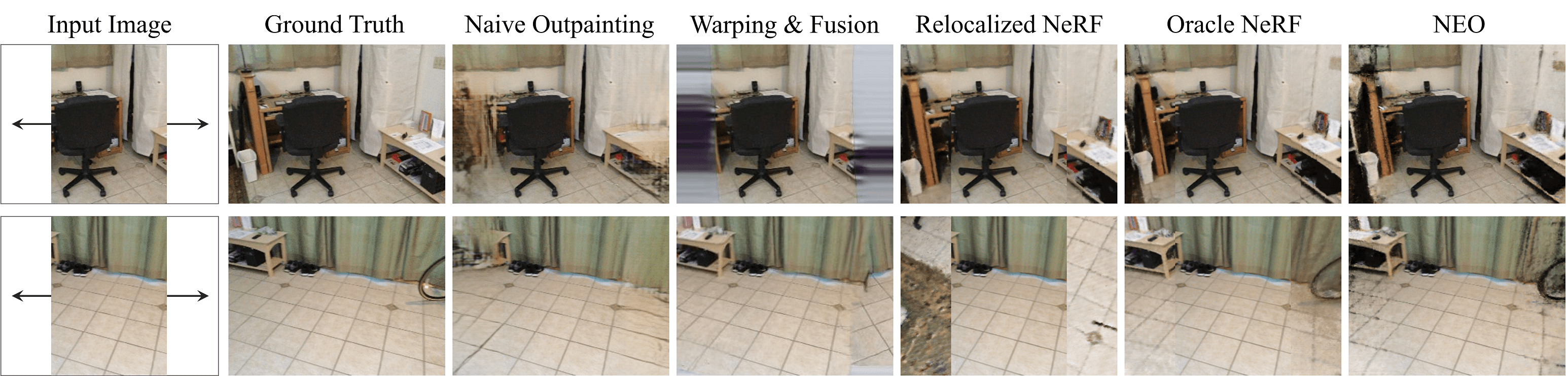} 
\caption{Qualitative results on \textbf{ScanNet} dataset.}
\label{fig:vis_result_scannet}
\end{figure*}

\subsection{Baseline Methods}
To demonstrate the effectiveness of our proposed NEO, we first introduce three baseline methods to address this problem, as we have discussed earlier. The first method is the ``naive outpainting'' mentioned in Sec. \ref{discussion1}, where an outpainting model is trained only using the original small-FOV images. The original image is resized larger and its cropped central part is used as the small-FOV input, while the original image is used as the corresponding large-FOV target. The second baseline approach, dubbed ``warping \& fusion'' (B2), goes through an image stitching pipeline. To be specific, for a testing image, we first retrieve the nearest images from the training set, then build correspondence between the testing image and the retrieved~(source) images. Finally, image warping is employed to get a larger FOV image by warping and fusing the contents from source images. The last baseline is called ``relocalized NeRF''. We first train a NeRF model on the training images. For a testing image, we relocalize its camera pose by employing camera relocalization methods. Finally we render a large-FOV image with the trained NeRF and the relocalized pose.
\subsection{Datasets and Metrics}
We first evaluate the FOV extrapolation performance on three scenes from three photorealistic datasets:
1) \textbf{Replica} dataset~\cite{straub2019replica} includes $18$ realistic indoor scenes. We adopt the first floor of the \texttt{apartment0} scene for evaluation. 
2) \textbf{Gibson} dataset~\cite{xia2018gibson} includes realistic scans of $572$ full buildings. We adopt the \texttt{Bonesteel} building to verify our method.
~3) Habitat-Matterport 3D (\textbf{HM3D}) dataset~\cite{ramakrishnan2021habitat} includes realistic scans of $1,000$ buildings. We adopt scene \texttt{00065} for evaluation. 
To verify our method on real scenes, we further demonstrate our method on \textbf{ScanNet}~\cite{dai2017scannet} which samples posed RGB images from $1513$ real indoor scans. We adopt \texttt{scene0000\_00} as our training set and \texttt{scene0000\_01} whose data is sampled at the same scene but with different trajectories as our testing set.

\par
For photorealistic datasets, we consider a simplified scenario: a robot with a fixed height and a fixed front camera navigates in an indoor environment. In such a setting, the camera has a constant height and can only rotate horizontally, so the motion of the camera has only 3 DoFs. For each scene, we use the Habitat environment~\cite{savva2019habitat} to render $1,000$ training images with $256 \times 256$ resolution and $90^{\circ}$ FOV from random camera poses at a fixed height of 1.5$m$.
We then render $2,000$ testing images with the same resolution and FOV, which contains 40 random walking paths at the same height with 50 images on each path. The target resolution during testing is $512 \times 512$ ($126.87^{\circ}$ FOV) by uniformly extrapolating in four directions. For ScanNet, we simulate a challenging but more realistic scenario to extrapolate an input central image with resolution $240 \times 160$ in horizonal directions~(left and right sides), whose target resolution is $240 \times 320$. We uniformly sample the original training trajectory by image IDs with an interval $10$, leading to $558$ training images. For testing, we randomly sample three pieces of continuous trajectory, each with $200$ images, leading to $600$ testing views.

Since we require the extrapolated regions to be consistent with the scene, we adopt PSNR, SSIM and LPIPS~\cite{zhang2018unreasonable} as the evaluation metrics for faithful FOV extrapolation.



\subsection{Implementation Details}
\noindent \textbf{Photorealistic Datasets.} For 
Replica, Gibson, and HM3D datasets,
we use MAT~\cite{DBLP:conf/cvpr/LiLZQWJ22} as our default outpainting model but remove its style manipulation module for deterministic extrapolation. For warping \& fusion baseline (B2), we employ the pipeline proposed in~\cite{DBLP:conf/iccv/LiuL0CH21} as our implementation. 
For camera relocalization, we first apply an image retrieval method NetVLAD~\cite{arandjelovic2016netvlad} to retrieve the nearest images from the training set w.r.t. a testing image, then run COLMAP~\cite{schonberger2016structure} which generalizes well to different environements to get the relative pose between the testing image and the retrieved training image. Finally we transform the relative pose to the absolute pose using the known training pose. For NeRF model, we employ DirectVoxGO~\cite{DBLP:conf/cvpr/0004SC22} which leverages 3D voxel representation to accelarate training and inference.

For new data generation, we first derive the walkable areas from the 2D floor plans of the datasets. We sample new camera poses on a 2D horizontal grid with a default interval of $0.05m$. At each position, we uniformly sample $72$ yaw angles for the horizonal rotation. As a result, we sampled about 1.63 million, 1.43 million, 1.56 million camera poses for Replica, Gibson, and HM3D, respectively. 
\smallskip
\par
\noindent \textbf{ScanNet.} For ScanNet~\cite{dai2017scannet}, we instead use LaMa~\cite{DBLP:conf/wacv/SuvorovLMRASKGP22} for outpainting, to satisfy our need on different resolution and aspect ratios. For NeRF model, we replace DirectVoxGO with a state-of-the-art NeRF method~\cite{dai2023hybrid} on ScanNet to achieve better rendering performance. Other baselines are evaluated in a similar manner as on the photorealistic ones. 

\par
For new data generation, we aim to generate 6-DoF novel poses whose distribution is similar to the training trajectory. Specifically, on $x$-$y$ plane, we sample new camera poses on a uniform 2D horizonal grid with an interval of $0.2m$, and ensure the sampled trajectories are roughly covered by the training set, \ie, the Euclidean distance from the new pose to the nearest training pose on $x$-$y$ plane should be limited within a threshold of $0.3m$. For horizonal rotation~(yaw angle), we sample from a uniform distribution whose upper and lower bounds come from the training poses. For other DoFs~(vertical translation, pitch and roll rotation), we empirically discover that the training poses roughly follow a Gaussian distribution, thus we sample from a Gaussian distribution whose mean and standard deviation are calculated from the training poses. Then we employ NeRF to render novel views using these poses. However, practically we found that a non-negligible portion of the rendered images on ScanNet are too blurry to provide useful information for training the outpainting model. To address this problem, we apply a blur detection algorithm based on Laplacian variance to compute the blurry scores 
for filtering out blurry images.
Eventually, we generated $0.75$ million outpainting-trainable images on ScanNet. 

\subsection{Results}
\smallskip
\noindent{\bf Quantitative Results.}
Table~\ref{table:quant_results} illustrates the extrapolation results of the proposed NEO approach and three baseline methods on four datasets. 
In addition, for validation purposes, the performance of ``oracle NeRF'' is reported, which employs the groundtruth camera pose of the test image for NeRF rendering. ``Oracle NeRF'' reflects the quality of the trained NeRF model and sets an upper bound for (B3) relocalized NeRF.
Since NEO learns the outpainting from NeRF-augmented images, its performance is expected to be lower than that of ``oracle NeRF''.
As shown in Table~\ref{table:quant_results}, NEO significantly outperforms the three baseline methods. As anticipated, ``oracle NeRF'' achieves the best results although it is not for practical use since groundtruth camera poses for testing images are unknown. Although the three baseline methods produce reasonably good results, they encounter non-trivial problems with faithfulness, which we will demonstrate in qualitative results below. 


\smallskip
\noindent{\bf Qualitative Results.}
We show some qualitative results on Replica, Gibson, and HM3D datasets in Fig.~\ref{fig:vis_result_synthetic}. ``Oracle NeRF'' learns a geometrically consistent 3D representation thus demonstrates appealing rendering in a coherent way on the three datasets. The areas extrapolated by NEO are also much more accurate and faithful to the scene compared to the three baselines. NEO sometimes 
suffers from slight misalignment around small objects~(\eg, the painting on the left wall on the third row of Fig.~\ref{fig:vis_result_synthetic}). The extrapolated regions of (B1) naive outpainting tend to be blurry, which is mainly caused by the limited number of training images. The extended areas by (B2) warping \& fusion are limited by the non-overlapping regions of neighboring images. As for (B3) relocalized NeRF, the input region (central part) always misaligns with the extrapolated regions due to evident errors in pose estimation. Comparison of Results on the ScanNet dataset, as shown in Fig.~\ref{fig:vis_result_scannet}, leads to similar observations. Surprisingly, we found that NEO can avoid some issues encountered by ``orcale NeRF'' and achieve better visual quality in some regions, such as the blurry floor region near the border of extrapolated image by ``oracle NeRF'' in the second row of Fig.~\ref{fig:vis_result_scannet}. We suspect the reason is that, NeRF is trained on sparse views from the original training set, thereby its rendering quality in specific areas may be dependent on the availability of informative, overlapping training images. 
In contrast, NEO trains the outpainting model on sufficient, dense novel views rendered from NeRF, so it is more capable of learning a 
semantically and geometrically coherent color field of the scene, effectively reducing the impact of insufficient information in some areas of the original NeRF.

\subsection{Discussions}

\smallskip
\noindent{\bf Pose Sampling.}
It is important in the NEO approach to cover as many views as possible in the scene in Step 2 of the training process. 
Thus, the distribution and number of sampled poses are crucial for training a highly effective outpainting model.
In this study, we vary the interval of the 2D grid to control the sampling density on the Replica dataset. As shown in Table~\ref{table:ablation}~(a), the generative performance naturally improves when increasing the pose sampling density. The improvement is significant (+1.30 in PSNR) when reducing the interval from $0.1m$ to $0.05m$, where the number of sampled poses increases from $0.4M$ to $1.6M$.

\begin{table}[t]
\small
\centering
\caption{Effect of (a) \textit{pose sampling density} and (b) \textit{FOV of training images} in NEO pipeline on Replica dataset.}
\label{table:ablation}
\begin{tabular}{c c c c c c} 
\toprule
{} & Interval & {\# Pose} & {FOV} & {PSNR $\uparrow$} & {SSIM $\uparrow$} \\ 
\midrule
\multirowcell{4}{(a)} & \textbf{0.05} & 1,629,864 & extended & 25.94 & 0.868 \\
{} & \textbf{0.10} & 406,224 & extended & 24.64 & 0.851 \\
{} & \textbf{0.20} & 101,304 & extended & 24.52 & 0.849 \\
{} & \textbf{0.50} & 16,056 & extended & 23.37 & 0.833 \\
{} & \textbf{1.00} & 3,744 & extended & 22.72 & 0.824 \\
\midrule
\multirowcell{2}{(b)} & 0.05 & 1,629,864 & \textbf{extended} & 25.94 & 0.868 \\
{} & 0.05 & 1,629,864 & \textbf{original} & 20.61 & 0.802 \\
\bottomrule
\end{tabular}
\vspace{-4mm}
\end{table}

\begin{figure}[t]
\setlength{\abovecaptionskip}{1mm}
\centering
\subfloat[FOV angles\label{subfig:fov1}]{
  \includegraphics[width=0.31\linewidth]{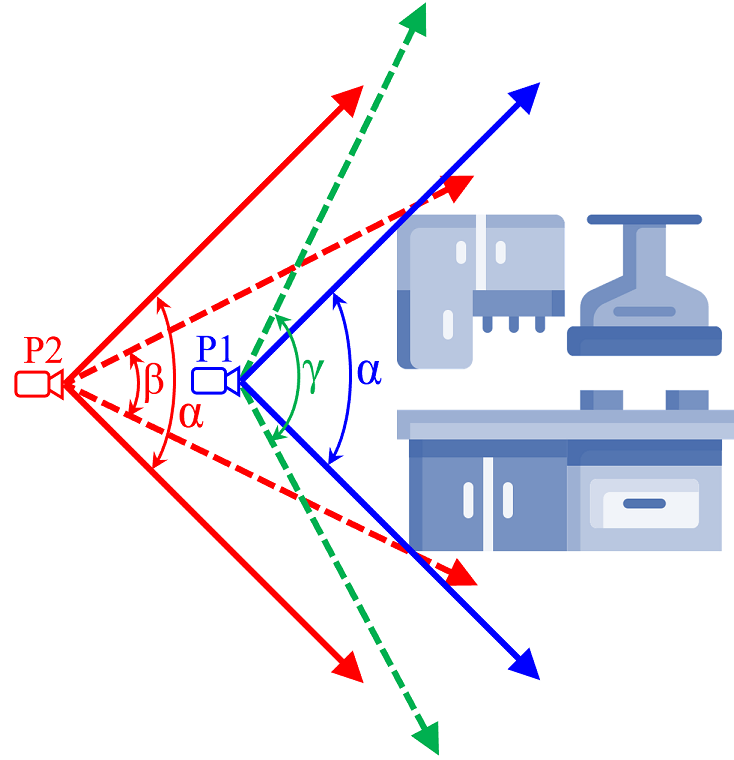}
}
\hspace{-2mm}
\subfloat[Image at Pose 1\label{subfig:fov2}]{
  \includegraphics[width=0.27\linewidth]{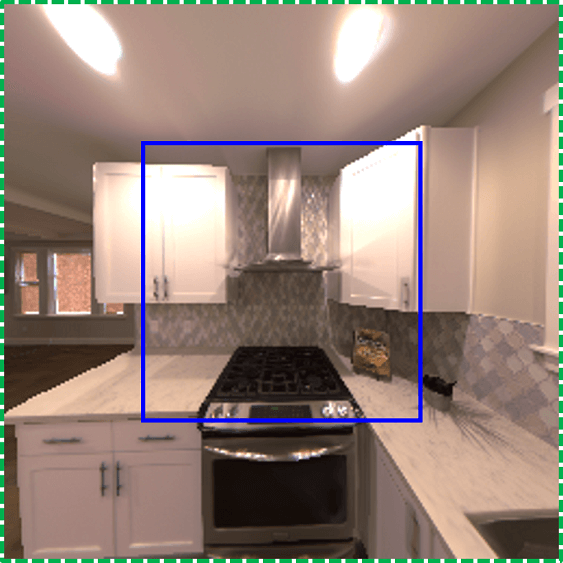}
}
\hspace{-2mm}
\subfloat[Image at Pose 2\label{subfig:fov3}]{
  \includegraphics[width=0.27\linewidth]{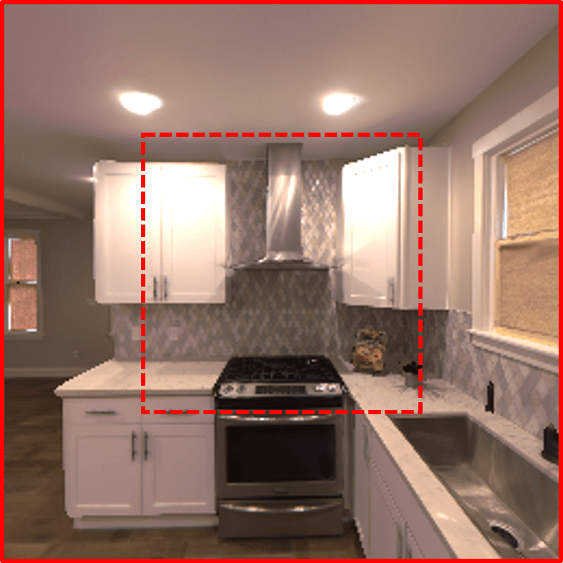}
}
\caption{FOV analysis. The discrepancy in the FOV between the (c) training image and (b) testing image may lead to false extrapolation behaviors.\label{fig:fov_analysis}}
\vspace{-10pt}
\end{figure}

\smallskip
\noindent{\bf FOV of Training Images.}
A key issue of training an outpainting model for FOV extrapolation is the consistency in FOV between training and testing images. Figure~\ref{fig:fov_analysis} demonstrates the FOV mismatch problem in the naive outpainting method. 
The solid red and blue arrows in Fig.~\ref{fig:fov_analysis}(a) represents the camera's inherent FOV $\alpha$. From a resized training image captured at Pose 2 (Fig.~\ref{fig:fov_analysis}(c)), naive outpainting learns to extrapolate the cropped FOV $\beta$ (dashed red) to $\alpha$. However, for a testing input image (solid blue) at Pose 1  (Fig.~\ref{fig:fov_analysis}(b)), the goal is to extrapolate the inherent FOV $\alpha$ to a larger FOV $\gamma$ (dashed green). 
Though the central parts (inputs) of Fig.~\ref{fig:fov_analysis}(b) and Fig.~\ref{fig:fov_analysis}(c) are similar, their extrapolated parts are totally different.
The FOV mismatch issue of naive outpainting can also be observed in the qualitative results (\eg, the painting on the second row in Fig.~\ref{fig:vis_result_synthetic}). To further examine the effect of FOV, we evaluate a variant of NEO, which uses the original-FOV synthetic images to train the same outpainting model. As seen in Table~\ref{table:ablation}~(b), the performance greatly decreases (-5.33 in PSNR), indicating the significance of training FOV.

\section{Limitations and Conclusions}
\label{sec:conclusion}

As an initial exploration of the faithful FOV extrapolation task, this paper focuses on tackling the problem for only static scenes. However, real-world navigation usually entails dynamic objects or people. Moreover, scenes are rarely static over time, \eg, furniture may be rearranged. This study can serve as a probe to more comprehensive research in this area. In the future, we envision to explore solutions that can accommodate the complexities of more realistic scenarios. One way may leverage dynamic NeRF~\cite{li2023dynibar} that better handles dynamic scenarios.

To conclude, in this paper, we formulate a new problem named {\it faithful image extrapolation} to increase FOV of a given image. It requires the expanded area to adhere to the real environment. To address this problem, inspired by the recent surge of NeRF-based rendering approaches, we propose a novel pipeline dubbed NEO, to train a NeRF-enhanced image outpainting model. Our key insight is to obtain sufficient and interpretable training data to aid the training of outpainting model from the novel views rendered by NeRF on a specific scene. Compared with competing baselines, our model has showcased superior generative performance. Our synthesized views are geometrically and semantically consistent with the 3D environment, thereby achieving faithful extrapolation that opens up potential applications such as AR-based navigation. 


\newpage
\bibliographystyle{IEEEtran}
\bibliography{IEEEfull}

\end{document}